\documentclass[letterpaper, 10 pt, conference]{ieeeconf}
\IEEEoverridecommandlockouts
\let\labelindent\relax
\usepackage{graphicx} \graphicspath{ {figures/} }
\usepackage{amsmath,amssymb,mathabx}
\usepackage{algorithm}
\usepackage{algorithmic}
\usepackage{acronym}
\usepackage{enumitem}
\usepackage{booktabs}
\usepackage{hyperref}
\usepackage{stfloats}
\usepackage{cuted}
\usepackage{balance}
\usepackage{xspace,setspace}
\usepackage[skip=3pt,font=small]{subcaption}
\usepackage[skip=3pt,font=small]{caption}
\usepackage[dvipsnames]{xcolor}
\usepackage[capitalise]{cleveref}
\usepackage{tabularx,colortbl,multirow,array,makecell}
\usepackage{overpic}
\usepackage{cite}
\usepackage{tikz}
\usepackage[]{mdframed}
\usepackage{soul}
\usepackage{anyfontsize}
\newcommand{\name}{\textit{CR-Solver}}

\makeatletter
\DeclareRobustCommand\onedot{\futurelet\@let@token\@onedot}
\def\@onedot{\ifx\@let@token.\else.\null\fi\xspace}

\makeatother

\crefname{algorithm}{alg.}{algs.}
\Crefname{algorithm}{Algorithm}{Algorithms}

\makeatletter
\def\BState{\State\hskip-\ALG@thistlm}
\makeatother

\makeatletter
\renewcommand{\paragraph}{%
  \@startsection{paragraph}{4}%
  {\z@}{0ex \@plus 0ex \@minus 0ex}{-1em}%
  {\hskip\parindent\normalfont\normalsize\bfseries}%
}
\makeatother

\crefname{algorithm}{alg.}{algs.}
\Crefname{algorithm}{Algorithm}{Algorithms}

\definecolor{gblue}{HTML}{4285F4}
\definecolor{gred}{HTML}{DB4437}

\acrodef{dof}[DoF]{Degree of Freedom}
\acrodef{vkc}[VKC]{Virtual Kinematic Chain}
\acrodef{tamp}[TAMP]{Task and Motion Planning}
\acrodef{pddl}[PDDL]{Planning Domain Definition Language}
\acrodef{rrt}[RRT]{Rapidly-exploring Random Tree}
\acrodef{ompl}[OMPL]{Open Motion Planning Library}
\acrodef{iws}[IWS]{Iterated Width Search}
\acrodef{bfs}[BFS]{Breadth First Search}
\acrodef{ai}[AI]{Artificial Intelligence}
\acrodef{vln}[VLN]{Vision-Language Navigation}
\acrodef{3dsg}[3DSG]{graph-based scene representation}
\acrodef{llm}[LLM]{Large Language Model}
\acrodef{pog}[PoG]{Planning on Graph}
\acrodef{epog}[EPoG]{Exploration and Planning on Graph}
\acrodef{ged}[GED]{Graph Edit Distance}

\definecolor{gblue}{HTML}{4285F4}
\definecolor{gred}{HTML}{DB4437}

\definecolor{custorange}{RGB}{255, 147, 30}
\definecolor{custblue}{RGB}{63, 167, 243}
\definecolor{custdarkblue}{RGB}{38, 99, 145}
\definecolor{custgrey}{RGB}{202, 202, 202}
\definecolor{custgreen}{RGB}{34, 139, 34}

\colorlet{lightorange}{custorange!20}

\colorlet{lightblue}{custblue!20}

\colorlet{lightgrey}{custgrey!40}

\title{\LARGE \bf \name{}: GPU-Accelerated Kinematics Solver for Tendon-driven Continuum Robots}

\author{Heqing Yang$^{1}$\quad{}Yang Yi$^{1}$\quad{}Linqing Zhong$^{1}$\quad{}Linjiang Huang$^{1\dagger}$\quad{}Si Liu$^{1\dagger}$
\thanks{This research is supported in part by the National
Natural Science Foundation of China (No. 62461160308,
U23B2010, 62576024), the Beijing Natural Science Foun
dation (No. L231011), the Fundamental Research Funds
for the Central Universities (No. 501RCQD2025141003),
BeiHang GanWei Project (No. 502GWXM2024141001)}
\thanks{$^\dagger$~Corresponding authors. $^{1}$~Beihang University.}%
}

\begin{document}

\maketitle
\thispagestyle{empty}
\pagestyle{empty}

\begin{abstract}
Continuum robots provide intrinsic compliance, high dexterity, and safe physical interaction, enabling navigation and manipulation in confined and unstructured environments. Despite recent advances in sensing and control, heightening the need for precise motion generation, most widely used planning libraries are grounded in rigid-body assumptions, creating a critical gap for fast and practical tools for continuum robots. To address this, we present \name{}, a two-stage, optimization-based solver for the motion generation of tendon-driven continuum robots. Our method unifies inverse kinematics, path following, and trajectory planning within a single constrained nonlinear optimization framework. Leveraging GPU-accelerated parallel optimization, \name{} delivers fast, accurate, and constraint-aware solutions. We validate our approach on three tasks, demonstrating significant speedups over traditional CPU-based solvers while achieving a consistently high success rate above 95\% and millimeter-level accuracy. The solver is implemented in pure \texttt{Python}, reducing the barrier to adoption and offering a practical, extensible foundation for continuum robots' high-performance motion planning. The source code is publicly available at \url{https://github.com/Noietch/CR-Solver}.
\end{abstract}

\setstretch{0.93}

\section{Introduction}

Continuum robots differ from conventional rigid-body manipulators by generating motion through deformation of a continuous backbone, which endows them with theoretically infinite degrees of freedom~\cite{russo2023continuum}. Their intrinsic structural compliance promotes safer human-robot interaction, while their slender morphologies enable navigation in confined, tortuous environments~\cite{sirohi2019design, abah2021multi}. These characteristics have motivated applications ranging from minimally invasive surgery to search-and-rescue~\cite{burgner2015continuum, kolachalama2020continuum}.

However, the inherent flexibility that enables the high dexterity of continuum robots also creates significant computational challenges. Their hyper-redundancy and highly nonlinear dynamics make motion generation computationally demanding. A rich ecosystem of motion planning tools exists for rigid-body robotics, spanning from classical planners like OMPL~\cite{sucan2012open} to modern GPU-accelerated solvers such as cuRobo~\cite{sundaralingam2023curobo} and PyRoki~\cite{kim2025pyroki}. Nevertheless, these established frameworks are fundamentally incompatible. They are grounded in URDF-based~\cite{tola2023understanding} rigid link abstractions defined by discrete joint coordinates. In contrast, the configuration spaces of continuum robots are described by continuous deformation fields, rendering the core modeling assumptions of these conventional frameworks invalid. Bridging this gap would require substantial architectural changes, highlighting the need for a purpose-built solution.

Within the continuum robotics community, existing strategies often mitigate this complexity by leveraging configuration-specific assumptions, such as the widely-used piecewise constant curvature (PCC) simplification~\cite{qiu2025actuator, kuntz2019planning}. While effective for certain robot morphologies, this approach sacrifices generality and scales poorly as system complexity increases~\cite{zhang2023cidgikc}. This reliance on simplifying models creates a significant bottleneck in the pursuit of a universal control framework. Thus, the fundamental incompatibility of rigid-body tools, coupled with the limited generality of existing specialized methods, strongly motivates the development of a new, scalable, and general-purpose solver tailored to the unique challenges of continuum robotics.

\begin{figure}[t]
\centering
\includegraphics[width=\linewidth]{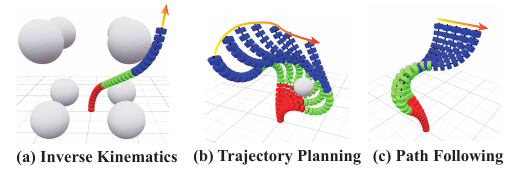}
\captionof{figure}{Our framework provides three core capabilities for continuum robots: (a) inverse kinematics that computes feasible configurations for desired end-effector poses while avoiding obstacles, (b) trajectory planning that generates collision-free trajectories between start and goal configurations, and (c) path following that enables precise tracking of predefined end-effector trajectories. All three functionalities leverage GPU-accelerated parallel optimization for efficient computation.}
\label{fig:overview}
\end{figure}

In this paper, we present \name, an efficient and unified framework for the inverse kinematics, trajectory planning, and path following of continuum robots, as illustrated in Fig.~\ref{fig:overview}. At its core is a GPU-accelerated solver built in pure \texttt{Python} on \texttt{JAX} \cite{jax2018github} with JIT compilation and automatic differentiation. To the best of our knowledge, it is the first solver in continuum robotics to leverage GPU acceleration. \name{} provides: 1)~\textbf{Configuration generality}, seamlessly supporting continuum robots with various scalable capabilities and numbers of segments; 2)~\textbf{Robust parallel optimization}, exploiting a two-stage optimization strategy that unleashes the capacity of GPU parallelism to optimize various solutions simultaneously, improving robustness to initialization and reducing susceptibility to local minima; and 3)~\textbf{Accessible tooling}, a concise and extensible codebase that lowers the barrier to adoption and research.

We conduct extensive experiments with curated comprehensive benchmarks spanning obstacle-free and cluttered settings, validating the performance over inverse kinematics, trajectory planning, and path following.
The results show that \name{} delivers both robustness and efficiency across all three tasks: it achieves nearly 100\% success in collision-aware inverse kinematics for continuum robots with varying segment counts, and attains millimeter-level accuracy in trajectory planning and path following. While substantially improving the success rate and accuracy, it also provides significant speed-ups over CPU-based baselines. We believe that the forthcoming open-source release of \name{} will offer the community a powerful and practical tool, accelerating innovation in continuum robotics.

\section{Related Works}

\subsection{Inverse Kinematics of Continuum Robots}
Forward kinematic modeling has been a central focus in continuum robotics research. Two high-fidelity paradigms, Cosserat rod formulations \cite{gazzola2018forward, zhang2019modeling, till2020dynamic} and finite element methods \cite{katzschmann2019dynamically}, capture the underlying continuum mechanics in detail. However, their reliance on solving complex partial differential equations makes them computationally prohibitive for real-time motion planning. The piecewise constant curvature (PCC) model \cite{webster2010design}, which represents continuum robots as sequences of constant-curvature arcs, has become the popular approach, offering an effective balance between computational speed and modeling accuracy.

Inverse kinematics solutions for PCC-based models broadly fall into two categories: analytical closed-form methods and numerical iterative approaches. Neppalli et al.~\cite{neppalli2009closed} reformulate the PCC model as a series of rigid links connected by spherical joints to derive closed-form solutions, although their approach notably omits end-effector orientation constraints. Garriga-Casanovas and Rodriguez~\cite{garriga2019kinematics} develop analytical solutions for both forward and inverse kinematics of continuum robots with bending and extension capabilities, yet their work does not address inextensible configurations. Numerical methods have gained prominence, with heuristic-based iterative approaches. FABRIK-C~\cite{zhang2018fabrikc} leverages a kinematic representation where each constant-curvature segment is replaced by two rigid links and a joint to simplify computation. However, it is limited to the target end-effector pose with 5-DoF. FABRIK-X~\cite{kolpashchikov2022fabrikx} extends FABRIK-C to handle variable curvature fixed-length segment robots, but lacks consideration of extensible robots. CIDGIKc \cite{zhang2023cidgikc} transforms the assumptions of the PCC model into a graph-based representation, which can reduce the number of iterations in numerical iteration. However, it only considers extensible robots, and its computation time increases rapidly with the number of robot segments. Qiu et al.~\cite{qiu2023efficient} demonstrate efficient inverse kinematics solving for three-segment continuum robots in obstacle-rich scenarios through model simplification, though their approach remains limited to this specific configuration. Therefore, the absence of a unified, scalable inverse kinematics framework motivates our GPU-parallelized approach that leverages sampling and optimization techniques for generalized solutions.

\subsection{Motion Planning of Continuum Robots}
Path following and trajectory planning are both subproblems of motion planning for continuum robots. Path following focuses on tracking a prescribed geometric path, while trajectory planning additionally time-parameterizes motion to satisfy kinematic, dynamic, and environmental constraints. Motion planning for continuum robots, analogous to rigid-body systems, encompasses sampling-based methods~\cite{kavraki2002probabilistic, lavalle2001randomized, karaman2011sampling} and optimization-based approaches~\cite{schulman2013finding, park2012itomp, ratliff2009chomp}. While sampling-based methods such as the RRT* algorithm have been adapted for continuum robots~\cite{luo2024efficient}, they often suffer from prohibitive computational costs. Qiu et al.~\cite{qiu2025actuator} propose a more efficient approach by computing multiple inverse kinematics solutions for each waypoint and employing dynamic programming to find optimal paths, however this method cannot accommodate extensible continuum robots. Optimization-based methods, exemplified by Lai et al.~\cite{lai2022constrained}, formulate motion planning as constrained optimization problems but remain susceptible to initialization and local minima convergence. Hybrid approaches combining both paradigms have shown promising results~\cite{kuntz2019planning},
but only validate on concentric tube robots. The aforementioned studies focus on modeling methods for specific robots and are not designed for robots based on more generalized modeling assumptions. Besides, these approaches are characterized by low computational efficiency. These limitations underscore the critical need for a unified, efficient, and generalizable motion planning framework for continuum robots.

\section{Preliminary} \label{sec:prelim}

\subsection{Kinematics Modeling of Continuum Robots} \label{sec:kinematics-modeling}

\begin{figure}[t]
    \centering
    \includegraphics[width=\linewidth]{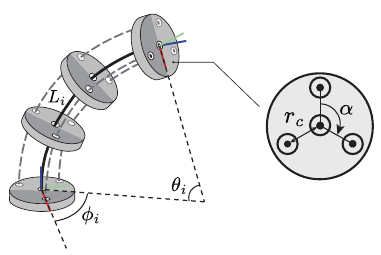}
    \caption{Kinematic model of a single segment in a tendon-driven continuum robot. Left: The segment structure showing the arc length $L_i$, bending angle $\theta_i$, and azimuth angle $\varphi_i$ with coordinate frames attached at each disk. Right: Cross-sectional view showing the tendon routing pattern with three tendons symmetrically placed at radius $r_c$ from the central backbone, with angular spacing $\alpha$ between adjacent tendons.}
    \label{fig:kinematics}
\end{figure}

We model continuum robots using the Piecewise Constant Curvature (PCC) model, which approximates the robot's backbone as a series of constant-curvature segments. For a continuum robot with $n$ segments, as shown in Fig.~\ref{fig:kinematics}, each segment $i$ is characterized by three parameters: the bending angle $\theta_i \in [0, \pi]$, the azimuth angle $\varphi_i \in [0, 2\pi]$, and the arc length $L_i \in \mathbb{R}^+$.

The forward kinematics can be calculated by chaining transformations of consecutive segment frames, where the transformation from frame $i-1$ to frame $i$ is represented by the homogeneous transformation matrix $\mathbf{T}_{i-1}^{i} \in \mathrm{SE}(3)$:
\begin{equation}
\mathbf{T}_{i-1}^{i} = \begin{bmatrix} \mathbf{R}_{i-1}^{i} & \mathbf{d}_{i-1}^{i} \\ \mathbf{0}^T & 1 \end{bmatrix},
\end{equation}
where $\mathbf{R}_{i-1}^{i} \in \mathrm{SO}(3)$ represents the rotation matrix and $\mathbf{d}_{i-1}^{i} \in \mathbb{R}^3$ denotes the translation vector.
The rotation component follows the composition of intrinsic rotations:
\begin{equation}
\mathbf{R}_{i-1}^{i} = \mathbf{Z}(\varphi_i)\mathbf{Y}(\theta_i)\mathbf{Z}(-\varphi_i),
\end{equation}
where $\mathbf{Z}$ and $\mathbf{Y}$ represent elementary rotations of
the z-axis and y-axis, respectively.
The translation vector is given by:
\begin{equation}
\mathbf{d}_{i-1}^{i} = r_i \begin{bmatrix} \cos\varphi_i(1-\cos\theta_i) \\ \sin\varphi_i(1-\cos\theta_i) \\ \sin\theta_i\end{bmatrix},
\end{equation}
where $r_i = L_i/\theta_i$ denotes the curvature radius of segment $i$. For the special case when $\theta_i = 0$, the segment becomes straight with $\mathbf{d}_{i-1}^{i} = [0, 0, L_i]^T$.

\subsection{Tendon-driven Actuation Model} \label{sec:tendon-driven-actuation-model}
For tendon-driven continuum robots, actuation is achieved through tendons routed parallel to the backbone of the robot in radius $r_c$ from the central axis. For segment $i$ with $m$ tendons uniformly distributed around the circumference, the $j$-th tendon ($j = 1, 2, \ldots, m$) is positioned at angular location $\alpha_j = (j-1) \cdot 2\pi/m$. The length of the $j$-th tendon in the segment $i$ is:

\begin{equation}
\ell_{i,j} = L_i - r_c \theta_i \cos(\alpha_j - \varphi_i),
\end{equation}
where the projection factor $\cos(\alpha_j - \varphi_i)$ determines the tendon's position relative to the bending plane of segment $i$. For a $n$-segment continuum robot with configuration $\mathbf{q}$, the total length of the $j$-th tendon spanning all segments is $\ell_j(\mathbf{q}) = \sum_{i=1}^{n} \ell_{i,j}$.

Collecting all tendons, we define the tendon-length mapping as the vector
\begin{equation}
\boldsymbol{\ell}(\mathbf{q}) = [\ell_1(\mathbf{q}),\, \ell_2(\mathbf{q}),\, \ldots,\, \ell_m(\mathbf{q})]^T \in \mathbb{R}^{m}.\label{eq:l_eq}
\end{equation}



\section{Methodology} \label{sec:method}

\begin{figure}[!t]
    \centering
    \includegraphics[width=\linewidth]{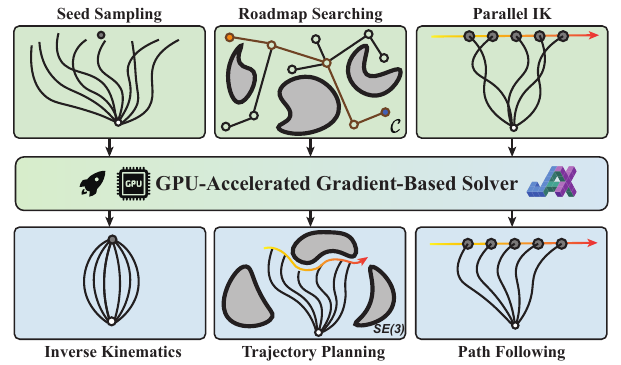}
    \caption{Framework of our \name. We employ a GPU-accelerated two-stage pipeline: Stage I performs massively parallel seed sampling, roadmap searching, and parallel IK to generate diverse feasible initializations; Stage II uses a batched gradient-based solver on a GPU to refine selected candidates, solving inverse kinematics, trajectory planning, and path following under a unified constrained optimization.}
    \label{fig:framework}
    \vspace{-1em}
\end{figure}

We formulate inverse kinematics, trajectory planning, and path following for continuum robots within a unified constrained nonlinear optimization framework, specifying both objectives and constraints:
\begin{align}
\min_{\mathbf{z}} \quad & f(\mathbf{z}) \\
\text{s.t.}\quad & c(\mathbf{z}) \le 0,\ \ h(\mathbf{z}) = 0.
\end{align}
Here, $\mathbf{z}$ denotes the unified decision variable, $f$ is the task objective, and the constraints are captured by the inequality set $c(\mathbf{z}) \le 0$ and the equality set $h(\mathbf{z}) = 0$. In what follows, we instantiate $f$, $c$, and $h$ to obtain the specific formulations for inverse kinematics, trajectory planning, and path following under this unified framework.

As shown in Fig.~\ref{fig:framework}, to ensure a robust and efficient solution for these tasks under the unified optimization formulation, we leverage the GPU's parallel computing capabilities by employing a two-stage optimization paradigm. This approach first utilizes massively parallel sampling and coarse optimization to generate high-quality initializations, exploring the solution space to identify promising candidates. Subsequently, a GPU-accelerated, gradient-based solver takes top-performing candidates and rapidly refines them, ensuring fast convergence to a feasible and precise solution.

\subsection{Task-specific Optimization Formulation}
\noindent \textbf{Inverse Kinematics (IK)} 
for a continuum robot with $n$ segments aims to find a configuration vector \(\mathbf{q} = [\theta_1, \varphi_1, L_1, \ldots, \theta_n, \varphi_n, L_n]^T \in \mathbb{R}^{3n}\) that realizes a desired end-effector pose \(\mathbf{T}_{\text{target}} \in \mathrm{SE}(3)\). Let \(\mathbf{T}_{\text{ee}}(\mathbf{q}) \in \mathrm{SE}(3)\) denote the end-effector's pose given by the forward kinematics mapping in Sec.~\ref {sec:kinematics-modeling}, we formulate IK as follows:
\begin{align}
\min_{\mathbf{q} \in \mathbb{R}^{3n}} \quad & \, \big\|\log\!\big(\mathbf{T}_{\text{ee}}^{-1}(\mathbf{q})\, \mathbf{T}_{\text{target}}\big)\big\|_2 
\; \label{eq:ik_opt} \\
\text{s.t.} \quad & \mathbf{q}^{\min} \leq \mathbf{q} \leq \mathbf{q}^{\max}, \label{eq:ik_limits} \\
& d_{\min}(\mathbf{q}) - \delta \geq 0 ,\label{eq:ik_coll}
\end{align}
where \(\delta>0\) is a safety margin and \(d_{\min}(\mathbf{q})\) denotes the minimum signed distance to any self or environmental contact. The bound constraint (Eq.~\ref{eq:ik_limits}) encodes material and actuation limits on bending, azimuth, and segment length, while the distance constraint (Eq.~\ref{eq:ik_coll}) enforces collision avoidance in a compact form.

\vspace{1mm}
\noindent \textbf{Trajectory Planning} aims to find a discrete trajectory \(\mathbf{Q} = [\mathbf{q}_0, \mathbf{q}_1, \ldots, \mathbf{q}_H] \in \mathbb{R}^{3n \times (H+1)}\) connecting given start and goal configurations as follows:
\begin{align}
\min_{\mathbf{Q}} \quad & \sum_{t=1}^{H} \big\| \mathbf{p}_{\text{ee}}(\mathbf{q}_t) - \mathbf{p}_{\text{ee}}(\mathbf{q}_{t-1}) \big\|_2 \label{eq:mp_length_only} \\
\text{s.t.} \quad & \mathbf{q}^{\min} \le \mathbf{q}_t \le \mathbf{q}^{\max}, \quad t=0,\ldots,H \\
& \mathbf{v}^{\min} \le \dot{\boldsymbol{\ell}}_t \le \mathbf{v}^{\max}, \quad t=1,\ldots,H \\
& \mathbf{a}^{\min} \le \ddot{\boldsymbol{\ell}}_t \le \mathbf{a}^{\max}, \quad t=2,\ldots,H \\
& d_{\min}(\mathbf{q}_t) - \delta \ge 0, \quad t=0,\ldots,H \\
& \mathbf{q}_0 = \mathbf{q}_{\text{start}}, \quad \mathbf{q}_H = \mathbf{q}_{\text{goal}}, \nonumber
\end{align}
where \(\mathbf{p}_{\text{ee}}(\mathbf{q}_{t}) \in \mathbb{R}^3\) is the end-effector position extracted from \(\mathbf{T}_{\text{ee}}(\mathbf{q}_{t})\), and \(\boldsymbol{\ell}(\mathbf{q}_{t}) \in \mathbb{R}^{n_{\ell}}\) denotes the tendon-length mapping defined in Sec.~\ref{sec:tendon-driven-actuation-model}.
Apart from the bound constraint and the distance constraint inherited from IK, we further restrict the tendon speed and acceleration, which can be calculated by finite differences as \(\dot{\boldsymbol{\ell}}_t = (\boldsymbol{\ell}(\mathbf{q}_t) - \boldsymbol{\ell}(\mathbf{q}_{t-1}))/\Delta t\), \(\ddot{\boldsymbol{\ell}}_t = (\boldsymbol{\ell}(\mathbf{q}_t) - 2\boldsymbol{\ell}(\mathbf{q}_{t-1}) + \boldsymbol{\ell}(\mathbf{q}_{t-2}))/\Delta t^2\), where \(\Delta t\) is the uniform time step.


\vspace{1mm}
\noindent \textbf{Path Following} represents a fundamental challenge in continuum robotics, where the robot must follow a predefined end-effector trajectory while maintaining structural integrity and avoiding obstacles. Our optimization-based framework naturally accommodates additional trajectory constraints beyond the collision avoidance and kinematic limits presented in the basic trajectory planning formulation. 

Concretely, given a desired end-effector trajectory $\mathbf{F}_{\text{ref}} = [\mathbf{T}_{\text{ee,0}}^{ref},\mathbf{T}_{\text{ee,1}}^{ref},...,\mathbf{T}_{\text{ee,}H}^{ref}]$, we can solve path following by replacing the objective function in Eq.~\ref{eq:mp_length_only} with:
\begin{equation}
\min_{\mathbf{Q}} \quad \sum_{t=0}^{H} \big\|\log\!\big(\mathbf{T}_{\text{ee}}^{-1}(\mathbf{q}_{t})\, \mathbf{T}_{\text{ee},t}^{ref}\big)\big\|_2\label{eq:constrained_mp},
\end{equation}
while maintaining all constraint conditions unchanged. 

\subsection{GPU-accelerated Two-stage Optimization}

\begin{algorithm}[!t]
\caption{Beam Search for Inverse Kinematics}
\label{alg:beam_search}
\begin{algorithmic}[1]
\REQUIRE Target pose $\mathbf{T}_{\text{target}}$, number of initial seeds $N_{\text{init}}$, number of final candidates $N_{\text{final}}$, iteration budgets $k_{\text{init}}$, $k_{\text{final}}$
\ENSURE Optimal configuration $\mathbf{q}^*$

\STATE {\color{blue}\textbf{// Generate initial seeds}}
\STATE $\mathbf{\Theta	}_{\text{init}} \leftarrow$ \textsc{SampleSeeds}($N_{\text{init}}$)

\STATE {\color{blue}\textbf{// Coarse optimization (all seeds in parallel)}}
\STATE $\mathbf{\Theta	}_{\text{opt}}, \mathbf{C}_{\text{opt}} \leftarrow$ \textsc{ParallelOptimize}($\mathbf{T}_{\text{target}}, \mathbf{\Theta	}_{\text{init}}$, $k_{\text{init}}$)

\STATE {\color{blue}\textbf{// Select best candidates}}
\STATE $\mathbf{\Theta}_{\text{best}} \leftarrow$ \textsc{SelectTop}($\mathbf{\Theta	}_{\text{opt}}$, $\mathbf{C}_{\text{opt}}$, $N_{\text{final}}$)

\STATE {\color{blue}\textbf{// Fine optimization (best candidates only)}}
\STATE $\mathbf{\Theta	}^*, \mathbf{C}^* \leftarrow$ \textsc{ParallelOptimize}($\mathbf{T}_{\text{target}}, \mathbf{\Theta	}_{\text{best}}$, $k_{\text{final}}$)

\RETURN $\mathbf{q}^* = \arg\min(\mathbf{C}^*)$
\end{algorithmic}
\end{algorithm}

The nonconvex nature of the aforementioned optimization problems poses a significant challenge, as it renders gradient-based methods highly sensitive to initialization and prone to convergence to sub-optimal local minima. Motivated by recent successes in GPU acceleration for solving rigid-body robots' problems, we implement our solver with the GPU's parallel computing capability to address the aforementioned optimization challenges. 

To facilitate GPU-friendly computation, we first convert all constraints into penalty terms with large weights and incorporate them into the objective function, thereby transforming the original constrained problem into an unconstrained multi-objective form:
\begin{align}
\min_{\mathbf{z}}~\operatorname{cost}(\mathbf{z}) := f(\mathbf{z}) + \rho_{\text{ineq}}\, \Pi_{\text{ineq}}\big(c(\mathbf{z})\big) + \rho_{\text{eq}}\, \Pi_{\text{eq}}\big(h(\mathbf{z})\big), \label{eq:penalized_form}
\end{align}
where $\text{cost}(\mathbf{z})$ denotes the aggregated objective after adding penalties, $\Pi_{\text{ineq}}$ penalizes only violated inequalities (zero when $c(\mathbf{z}) \le 0$), and $\Pi_{\text{eq}}$ penalizes equality residuals. 


To fully unleash the advantage of GPU in parallel computing and mitigate the risk of converging to local minima, we propose a \emph{two-stage} optimization strategy. 
\emph{Exploration stage} leverages the GPU's parallel computing capability to efficiently explore the solution space through task-specific strategies, such as massively parallel sampling or distributed optimization across multiple initializations. \emph{Refinement stage} selects the most promising candidates from the first stage and refines them using a GPU-accelerated gradient-based solver.

In the following, we proceed with IK, trajectory planning, and path following of continuum robots in accordance with this two-stage optimization method.


\vspace{1mm}
\noindent \textbf{Inverse Kinematics.}
Algorithm~\ref{alg:beam_search} casts IK as a GPU-accelerated beam search that follows the two-stage template, where the matrix $\mathbf{\Theta}$ contains candidate configuration vectors and $\mathbf{C}$ represents their costs, evaluated with Eq.~\ref{eq:penalized_form}. Specifically, \textsc{SampleSeeds} draws $N_{\text{init}}$ initial configurations within the joint bounds (Eq.~\ref{eq:ik_limits}) discarding any that immediately violate collision checks (Eq.~\ref{eq:ik_coll}). We then run $k_{\text{init}}$ steps on all $N_{\text{init}}$ seeds in parallel to quickly scout the configuration space. After these coarse steps, \textsc{SelectTop} ranks by Eq.~\ref{eq:ik_opt}, keeps the best $N_{\text{final}}$, and breaks near-duplicate ties in favor of lower collision penalties to encourage diversity. The selected candidates are further refined for $k_{\text{final}}$ iterations, and the batched solver \textsc{ParallelOptimize} executes every sub-problem concurrently on the GPU. The final answer selects the configuration with the minimal cost, i.e., $\mathbf{q}^* = \arg\min (\mathbf{C}^*)$.



\vspace{1mm}

\begin{algorithm}[!t]
\caption{GPU-Parallelized Trajectory Planning}
\label{alg:prm_planning}
\begin{algorithmic}[1]
\REQUIRE Start configuration $\mathbf{q}_{\text{start}}$, goal configuration $\mathbf{q}_{\text{goal}}$, roadmap parameters $(N, k)$, target horizon $H$
\ENSURE Optimized fixed-length trajectory $\mathbf{Q}^* = [\mathbf{q}_0, \ldots, \mathbf{q}_H]$ with $H{+}1$ waypoints
\STATE {\color{blue}\textbf{// Offline: Build roadmap (once)}}
\STATE $\mathcal{V} \leftarrow$ \textsc{ParallelSample}($N$)
\STATE $\mathcal{E} \leftarrow$ \textsc{ParallelConnect}($\mathcal{V}$, $k$)
\STATE $\mathcal{G} \leftarrow (\mathcal{V}, \mathcal{E})$

\STATE {\color{blue}\textbf{// Online: Query processing}}
\STATE $\mathbf{v}_s \leftarrow$ \textsc{ConnectToRoadmap}($\mathbf{q}_{\text{start}}$, $\mathcal{V}$)
\STATE $\mathbf{v}_g \leftarrow$ \textsc{ConnectToRoadmap}($\mathbf{q}_{\text{goal}}$, $\mathcal{V}$)
\STATE $\mathbf{Q}_{\text{path}} \leftarrow$ \textsc{Dijkstra}($\mathbf{v}_s$, $\mathbf{v}_g$, $\mathcal{G}$)
\STATE $\mathbf{Q}^{\prime} \leftarrow$ \textsc{ResampleLinear}($\mathbf{Q}_{\text{path}}$, $H$)
\STATE $\mathbf{Q}^* \leftarrow$ \textsc{ParallelOptimize}($\mathbf{Q}^{\prime}, \mathbf{q}_{\text{start}}, \mathbf{q}_{\text{goal}}$)
\RETURN $\mathbf{Q}^*$
\end{algorithmic}
\end{algorithm}

\noindent \textbf{Trajectory Planning.}
Algorithm~\ref{alg:prm_planning} adopts a hybrid strategy coupling a Probabilistic Roadmap (PRM)~\cite{kavraki2002probabilistic} global explorer with a local trajectory optimizer. The roadmap is built \emph{offline} once: \textsc{ParallelSample} generates $N$ collision-free configurations as nodes and \textsc{ParallelConnect} forms edges among their k-nearest neighbors in parallel to obtain the graph $\mathcal{G}=(\mathcal{V},\mathcal{E})$. 
For an \emph{online} query process, the exploration stage attaches the start and goal configurations to the roadmap via \textsc{ConnectToRoadmap} and then runs Dijkstra~\cite{dijkstra2022note} on $\mathcal{G}$ to return a discrete path. Because sampling-based paths have nonuniform waypoint spacing (implicit, variable time steps), we linearly interpolate this path in the configuration space using \textsc{ResampleLinear} to obtain a fixed horizon of $H{+}1$ waypoints, yielding the initial trajectory $\mathbf{Q}'$. The second stage refines $\mathbf{Q}'$ with the \textsc{ParallelOptimize} solver to enforce kinodynamic and collision constraints while minimizing the overall cost, producing the final trajectory $\mathbf{Q}^*$.

\vspace{1mm}
\noindent \textbf{Path Following.} In the exploration stage, we run parallel IK at each reference waypoint and assemble the resulting configurations into the initial trajectory. This ensures the end-effector starts on the reference path and is kinematically feasible, which improves robustness and accelerates convergence. In the refinement stage, we jointly optimize all waypoints with the GPU-accelerated solver to minimize the tracking objective in Eq.~\ref{eq:constrained_mp} while enforcing smoothness and kinodynamic penalties, yielding a collision-free, dynamically feasible trajectory.

\subsection{Implementation Details}

\noindent \textbf{GPU Acceleration.}
The entire system is implemented in pure \texttt{Python} using \texttt{JAX} and a GPU-accelerated least-squares backend, which provides automatic differentiation and accelerates computation, ensuring both high performance and ease of reproduction and extension. We execute batched trust-region Levenberg-Marquardt (LM)~\cite{more2006levenberg} solves with dense Cholesky linear algebra. The multi-start pipeline for IK runs fully in parallel across all initial seeds. For trajectory planning, key operations like roadmap node sampling and edge connectivity checks are also parallelized.

\vspace{1mm}
\noindent \textbf{Obstacle Convex Decomposition.} To effectively handle non-convex obstacles, we employ convex decomposition \cite{wei2022approximate}. This method approximates each non-convex geometry as a union of convex components. Based on this representation, distance computations can be performed as efficient, batched queries. Furthermore, it yields stable and piecewise-smooth penetration gradients, which improve optimizer convergence. All convex components can be precomputed and reused across experiments. The proposed scheme integrates seamlessly into our vectorized, Just-In-Time compiled pipeline to maintain real-time performance in cluttered environments.


\section{Experiments} \label{sec:sim}

\subsection{Environment Details}
All experiments were conducted on a workstation equipped with dual Intel Xeon Platinum 8480+ processors (3.8 GHz) and an NVIDIA GeForce RTX 4090 GPU (24 GB GDDR6X), running the Ubuntu 22.04 operating system with CUDA 12.4. Additionally, MATLAB R2024b was utilized for the re-implementation of specific algorithms.

\subsection{Inverse Kinematics}
For inverse kinematics, we evaluate algorithms on continuum robots composed of 3 and 4 segments. Each segment has a length of 1.0m and a radius of 0.1m. For non-extendable robots, the configuration parameters are bounded by $\theta_i \in [0, \pi]$ and $\varphi_i \in [0, 2\pi]$ with fixed length $L_i = 1.0$m. For extendable robots, we additionally optimize segment lengths $L_i \in [0.5, 1.5]$m. Each segment is discretized with 10 points for forward kinematics computation and collision detection. Our beam search algorithm begins with 128 initial seeds. The top 8 candidates from the initial stage are then selected for final optimization. The optimization process for each target pose is constrained by a total budget of 200 iterations.

The pose error is measured using the $\mathrm{SE}(3)$ logarithmic distance:
\begin{equation}\label{error_eq}
e = \big\|\log\!\big(\mathbf{T}_{\text{ee}}^{-1}(\mathbf{q})\, \mathbf{T}_{\text{target}}\big)\big\|_2,
\end{equation}
where $\mathbf{T}_{\text{ee}}$ is the computed end-effector pose and $\mathbf{T}_{\text{target}}$ is the target pose. Success rate is defined as the percentage of solutions that simultaneously satisfy: (1) pose accuracy with error $e < 0.01$, (2) configuration limit constraints, and (3) collision-free configurations when obstacles are present.

We evaluate our \name{} under two scenarios: basic IK without obstacles and collision-aware IK. While CPU-based baseline methods process targets sequentially, \name{} processes all 1000 targets in a single batch on both CPU and GPU. For comparison, we reimplement Micsolver~\cite{qiu2023efficient}, which represents the current state-of-the-art solution but is limited to 3-segment non-extendable robots, as well as CIDGIKc~\cite{zhang2023cidgikc} for extendable robot configurations. 

\subsubsection{Inverse Kinematics without Obstacles}

\begin{table}[!t]
\fontsize{8.5pt}{9pt}\selectfont
\setlength{\tabcolsep}{2pt}
\centering
\caption{Performance comparison for inverse kinematics without obstacles. * indicates the method without beam search. SR denotes success rate.}
\label{tab:ik_results}
\begin{tabular}{lcccc}
\toprule
\textbf{Method} & \textbf{Device} & \textbf{SR} (\%) & \textbf{Total Time} (s) & \textbf{Pose Error}\\
\midrule
Derivative-free~\cite{qiu2023efficient} & CPU & 57.70 & 176.93 & 9.94 $\times 10^{-3}$\\
Newton-Raphson~\cite{qiu2023efficient} & CPU & 91.90 & 15.82 & 2.91 $\times 10^{-3}$\\
Micsolver~\cite{qiu2023efficient} & CPU & 100.0 & 3.48 & 3.21 $\times 10^{-3}$\\
\midrule
\name{} & CPU & 100.00 & 404.29 & 3.74 $\times 10^{-4}$\\
\name{}* & GPU & 44.20 & \textbf{0.14} & 1.16 $\times 10^{-3}$\\
\textbf{\name{}} & \textbf{GPU} & \textbf{100.00} & 0.46 & \textbf{3.74 $\times$}$\mathbf{10^{-4}}$\\
\bottomrule
\end{tabular}%
\end{table}

\begin{figure}[t]
    \centering
    \includegraphics[width=\linewidth]{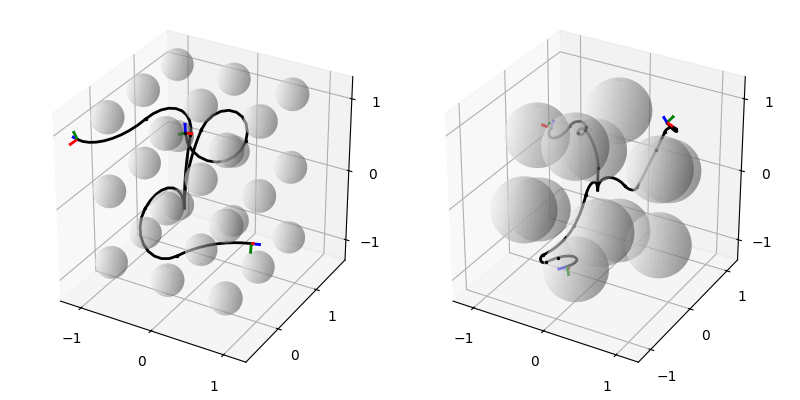}
    \caption{Inverse kinematics solutions for non-extendable (left) and extendable (right) robot configurations, demonstrating the solver's capability.}
    \label{fig:ik_examples}
    \vspace{-1.em}
\end{figure}

We evaluate IK performance on a benchmark of 1000 uniformly sampled target poses. Table~\ref{tab:ik_results} compares our method with representative baselines in the obstacle-free setting. Our approach attains a 100\% success rate and completes the benchmark in 0.46s. This runtime is over 7.5 times faster than Micsolver (3.48 s), indicating significant gains in computational efficiency. The ablation study on \name{}* (without beam search) highlights its critical role in ensuring solution reliability. Although removing this component reduces the average runtime per attempt, the success rate drops sharply. This demonstrates that beam search is essential for providing high-quality initial guesses that reliably lead the optimizer to a valid solution. Finally, the CPU-to-GPU comparison underscores the benefits of parallelism for batched IK, with GPU execution reducing the computation time from 404.29s to 0.46s without compromising solution quality.

\subsubsection{Inverse Kinematics with Obstacles}

For collision-aware inverse kinematics evaluation, we construct a structured obstacle environment to test the solver's capability in a challenging, cluttered scenario. The obstacle configuration consists of spherical obstacles arranged in a $3\times3\times3$ lattice pattern. Each sphere has a radius of 0.2m, positioned with inter-obstacle spacing of 0.8m in both $x$ and $y$ directions, and 1.0m spacing in the $z$ direction. This configuration results in a lattice of 27 spherical obstacles, creating a complex navigation space for the continuum robot. To ensure a fair evaluation, we sample 100 collision-free target poses. Example results of the inverse kinematics solver for non-extendable and extendable robots across different robot configurations are presented in Fig.~\ref{fig:ik_examples}. 

\begin{table}[!t]
\fontsize{9pt}{10pt}\selectfont
\setlength{\tabcolsep}{3pt}
\centering
\caption{Performance comparison for inverse kinematics with obstacles across different robot configurations. SR denotes success rate.}
\label{tab:collision_ik_results}
\begin{tabular}{lcccc}
\toprule
\textbf{Method} & \textbf{Segments} & \textbf{SR} (\%) & \textbf{Time} (s) & \textbf{Pose Error} \\
\midrule
\multicolumn{5}{c}{\textit{Non-extendable robots}} \\
\midrule
Micsolver\cite{qiu2023efficient} & 3 & 100.00 & 2.35 & 3.36 $\times$$10^{-3}$\\
\textbf{\name{}} & \textbf{3} & \textbf{100.00} & \textbf{0.30} & \textbf{1.00 $\times$}$\mathbf{10^{-4}}$\\
\midrule
\multicolumn{5}{c}{\textit{Extendable robots}} \\
\midrule
CIDGIKc \cite{zhang2023cidgikc} & 4 & 100.00 & 269.20 & 9.24 $\times 10^{-4}$\\
\textbf{\name{}} & \textbf{4} & \textbf{100.00} & \textbf{0.323} & \textbf{3.30 $\times$}$\mathbf{10^{-5}}$\\
\bottomrule
\end{tabular}

\end{table}

Table~\ref{tab:collision_ik_results} highlights the high performance of \name{} in collision-aware scenarios. For non-extendable robots, \name{} achieves a 100\% success rate and is approximately 7.8 times faster than Micsolver (0.30 s vs. 2.35 s). For extendable robots, \name{} maintains sub-second performance, whereas CIDGIKc requires 269.20 s for the same task, revealing a performance gap of over 800-fold in computational efficiency. The consistent sub-second runtime across all configurations demonstrates \name{}'s scalability and suitability for real-time applications.

\begin{figure}[t]
    \centering
    \includegraphics[width=1.0\linewidth]{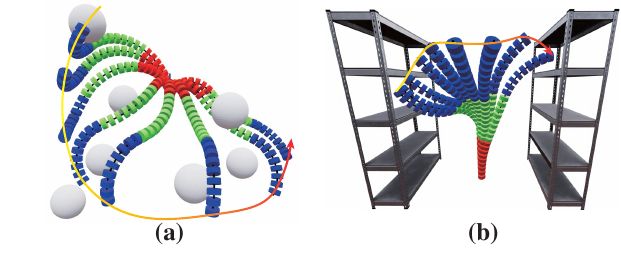}
    \caption{Trajectory planning in cluttered environments. Colored robot configurations depict a single trajectory, visualized every 13 timesteps; the orange curve with an arrow is the corresponding end-effector trace. (a) Random scene with spherical obstacles. (b) Shelf environment, where the trajectory traverses a narrow aisle while satisfying joint-limit and collision-avoidance constraints.}
    \label{fig:mp_examples}
    \vspace{-0.5em}
\end{figure}

\subsection{Trajectory Planning}

For the evaluation of trajectory planning, we generate a benchmark dataset comprising 296 start-goal configuration pairs distributed across multiple distinct 3D environments, including various shelf scenes, cabinet scenes, and randomly generated scenes (see Fig.~\ref{fig:mp_examples}). Randomly generated environment is a $3 \times 3 \times 3$\,m workspace centered at the origin; obstacles include structured shelf/cabinet layouts as well as randomly generated spherical clutter (27 spheres with radii sampled from $[0.15, 0.30]$\,m). To increase the likelihood that each planning problem is meaningful and feasible, collision-free start and goal configurations are selected such that the distance between their end-effector positions is at least $0.1 \times L_{\text{robot}}$, where $L_{\text{robot}}$ is the total robot length. This process yields a comprehensive benchmark for assessing planning performance in cluttered environments.

Here, we compare the trajectory planning performance of two different initialization strategies: a linear method based on direct interpolation between the start and goal configurations, and the heuristic method from Algorithm~\ref{alg:prm_planning}. The performance metrics included success rate (\%), trajectory length (m), and planning time (s). 

\begin{figure}[t]
    \centering
    \includegraphics[width=0.9\linewidth]{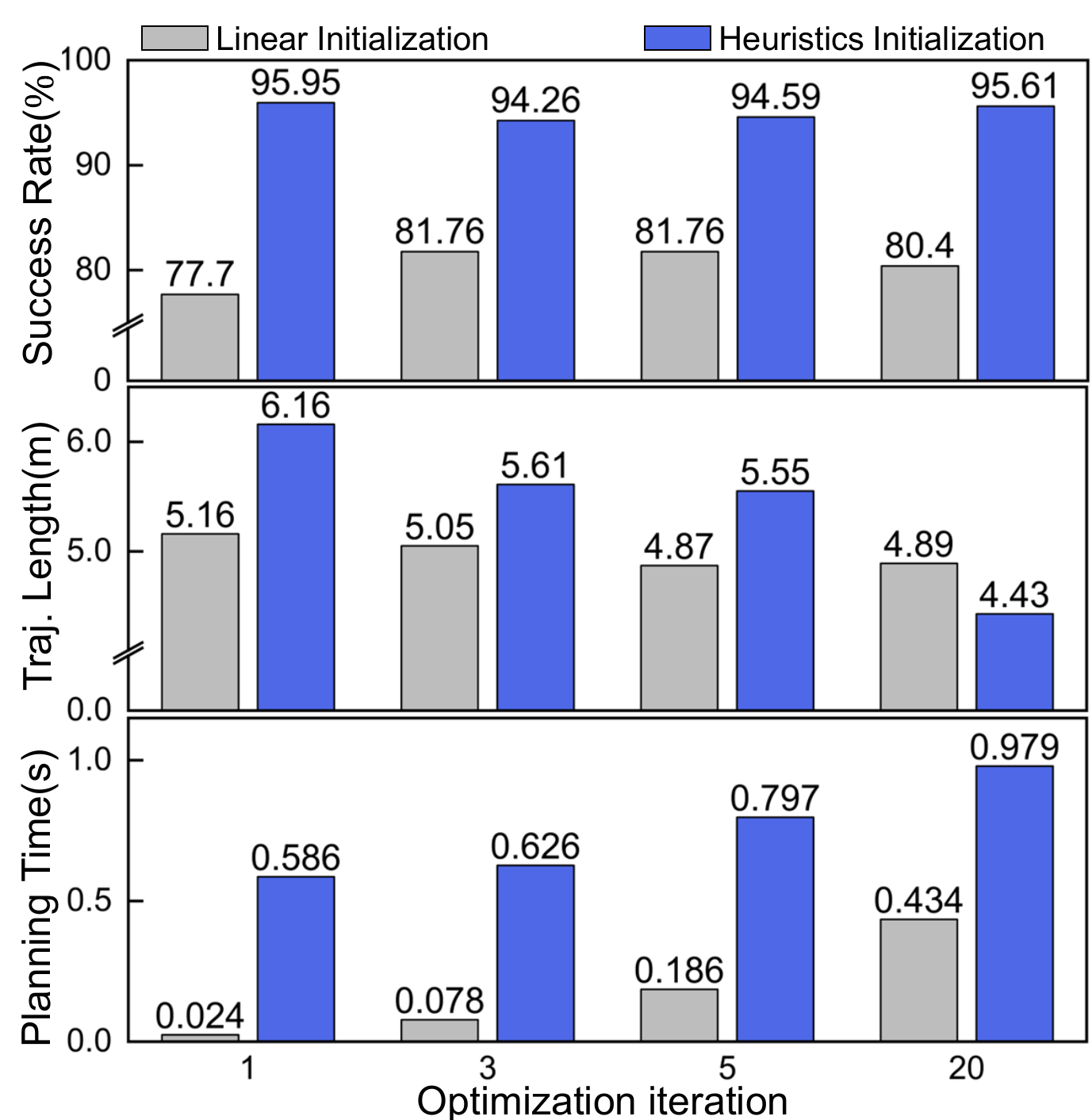}
    \caption{Ablation study on trajectory planning with different initialization strategies.}
    \label{fig:tp_result}
    \vspace{-1.em}
\end{figure}

\begin{figure*}[h]
    \centering
    \includegraphics[width=0.98\linewidth]{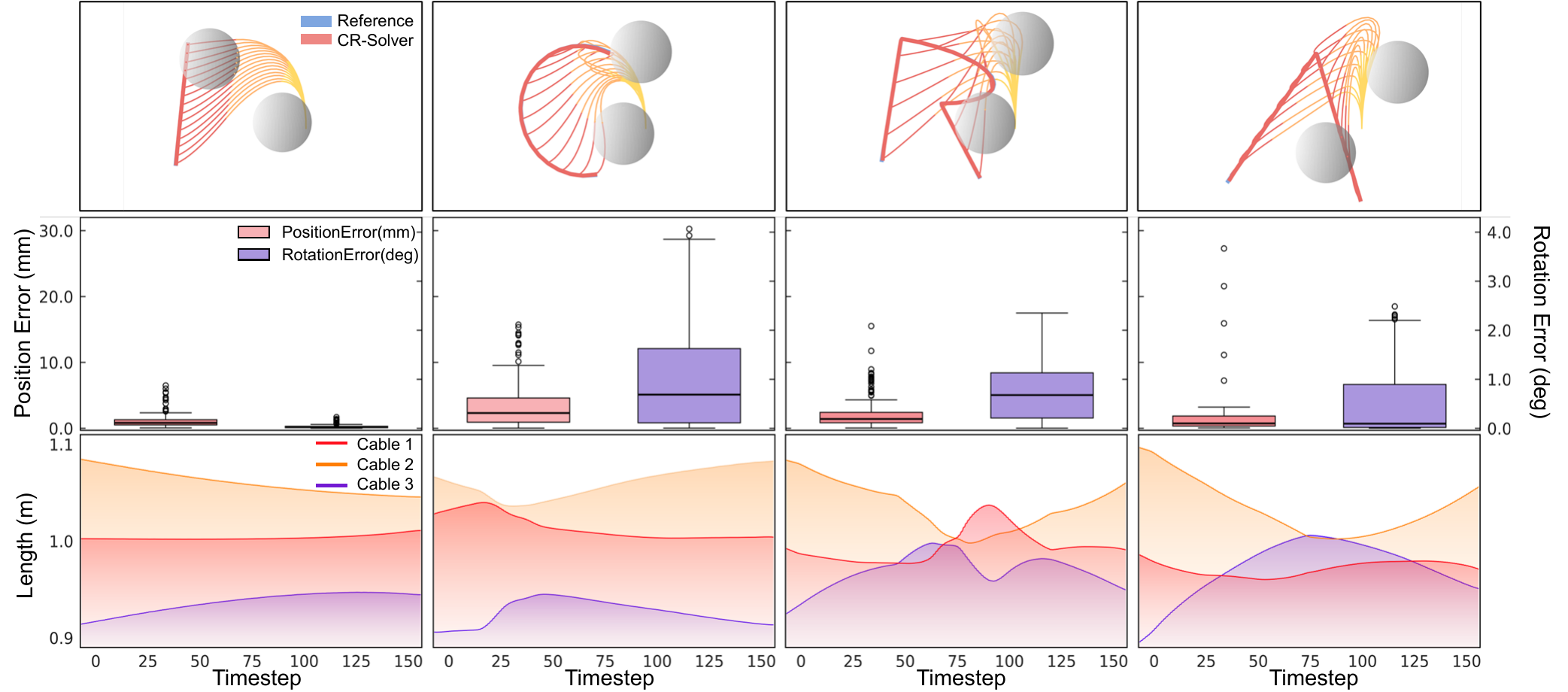}
    \caption{The results of tracking the ICRA-shaped paths. Top row: the robot traces the letters I, C, R, and A while avoiding obstacles, showing representative configurations along each stroke. Middle row: boxplots of tracking errors for each letter trajectory, including position error (mm, left axis) and rotation error (deg, right axis). Bottom row: evolution of the three driving cable lengths of the first segment during tracking, demonstrating smooth and continuous actuation.}
    \label{fig:constrained_mp}
    \vspace{-1.0em}
\end{figure*}

Given the scarcity of trajectory planning methods based on the PCC model, our experiment focuses on validating the effectiveness of our proposed initialization strategy. The results are presented in Fig.~\ref{fig:tp_result}. Incorporating the sampling-based heuristic initializer increases the mean success rate by approximately 15\%, accelerating convergence and reducing susceptibility to local minima in cluttered scenes. With a limited iteration budget, linear interpolation may yield shorter paths for simpler planning problems where the direct path is largely collision-free. As the number of optimization iterations increases, the heuristic initializer reveals a higher-quality search region, and the path length decreases steadily, surpassing linear initialization at approximately 20 iterations. The heuristic search introduces additional overhead, resulting in an end-to-end planning time of 0.97\,s compared to 0.45\,s with linear initialization. Inspired by previous RRT trajectory planning method~\cite{luo2024efficient}, we also implemented an RRT planner for comparison, which required 3.68\,s to achieve a lower success rate of 83.78\%. We also run our method on CPU, which requires 1.09\,s for three iterations, whereas the GPU-parallel variant reduces the search time by 43\% to 0.626\,s. As the optimization iteration budget grows, the total solve time increases accordingly. At 20 iterations, the heuristic initializer already surpasses the linear baseline in path quality while incurring the expected runtime increase. Despite this overhead, the heuristic initializer achieves a significantly higher success rate. Furthermore, it finds shorter trajectories in cluttered environments even with a modest iteration budget.

\subsection{Path Following}
We evaluate our path following algorithm on a 3-segment continuum robot in the environment with obstacles. Three distinct paths are designed to test different aspects of the path following algorithm: 1) a square path with sharp corners, 2) a sinusoidal path with smooth continuous motion, and 3) an ICRA-shaped path forming complex letter patterns. Each path consists of 150 waypoints with prescribed end-effector positions and orientations. The robot must track these waypoints while maintaining collision-free configurations and respecting joint limits. We compare our \name{} against Micsolver, evaluating position error (mm), rotation error (degrees), and computational time per path (s).

Table~\ref{tab:constrained_mp_results} presents the comparative performance between \name{} and Micsolver on three trajectory-following tasks. Our method, \name{}, achieves millimeter-level position accuracy across all trajectories, while Micsolver exhibits significantly larger position errors, particularly struggling with the sinusoidal trajectory, where smooth continuous motions pose challenges for its discrete optimization approach. Both methods maintain comparable rotation accuracy, demonstrating that orientation tracking is well handled by both approaches, though \name{} achieves this alongside its superior position tracking performance. Fig.~\ref{fig:constrained_mp} demonstrates the ability of our \name{} to execute complex trajectories with high precision and smooth control. 


\begin{table}[!t]
\fontsize{8pt}{9pt}\selectfont
\setlength{\tabcolsep}{1.5pt}
\centering
\caption{Path following performance. Pos. Err. and Rot. Err. Denote the position error and rotation error, respectively.}
\label{tab:constrained_mp_results}
\begin{tabular}{llccc}
\toprule
\textbf{Method} & \textbf{Shape} & \textbf{Pos. Err.} (mm) & \textbf{Rot. Err.} (deg) & \textbf{Time} (s)\\
\midrule
 \multirow{3}{*}{Micsolver\cite{qiu2023efficient}}&Square & 54.20 & 0.41 & 0.440\\
 &Sinusoidal & 168.01 & 0.03 & 0.559\\
 &ICRA-Shaped & 60.72 & 0.34 & 0.445\\
 \midrule
 \multirow{3}{*}{\textbf{\name{}}}&Square & 3.03 & 0.47 & 0.115\\
 &Sinusoidal & 3.76 & 0.25 & 0.080\\
 &ICRA-Shaped & 3.63 & 0.57 & 0.122\\
\bottomrule
\end{tabular}
\vspace{-1em}
\end{table}

In terms of computational efficiency, \name{} is 3 to 7 times faster than Micsolver, requiring less than 125 ms to complete each trajectory. In contrast, Micsolver requires substantially longer computation times, limiting its applicability in time-critical scenarios. The consistent performance of \name{} across diverse trajectory types, from sharp corners to smooth curves to complex patterns, validates its robustness and versatility as a general-purpose solution for path following in cluttered environments.

\section{Limitation} \label{sec:limit}

\noindent While our approach demonstrates strong performance across tasks, it has three main limitations that motivate future work:

\vspace{0.5em}
\noindent 1) Constraint handling via penalty reformulation. We handle constraints by reformulating the optimization problem into a penalized, unconstrained least-squares format to exploit GPU parallelism. This can handle most common hard constraints, but it does not guarantee adherence to all hard constraints.

\vspace{0.5em}
\noindent 2) Performance trade-off of a pure \texttt{Python}/\texttt{JAX} stack. Our modular \texttt{Python}/\texttt{JAX} implementation eases adoption and extension, but does not reach the peak throughput of hand-optimized C++/CUDA, reflecting a conscious trade-off between development velocity and absolute speed.

\vspace{0.5em}
\noindent 3) Modeling scope. We focus on PCC-based tendon-driven continuum robots. Extending this framework to alternative models (e.g., Cosserat rods, discrete elastic rods, or mechanism-specific parameterizations) would require additional development and validation.

\section{Conclusion} \label{sec:conclusion}

We have presented \name, an easy-to-use GPU-accelerated framework for inverse kinematics, trajectory planning, and path following of continuum robots, built on a \texttt{JAX}-based solver. \name{} supports diverse continuum robot configurations, exploits massive GPU parallelism for robust solution exploration, and offers accessible, extensible tooling. Extensive experiments show that \name{} demonstrates robustness and high efficiency in solving the kinematics problems of continuum robots. Although limitations exist, we aim to address them together with the community through open collaboration. We will release code and examples to encourage adoption and reuse, and we look forward to community contributions and applications in perception-in-the-loop control, large-scale design exploration, and real-time planning on embedded GPUs.

\bibliographystyle{ieeetr}
\bibliography{IEEEfull}

\end{document}